\documentclass[letterpaper, 10 pt, conference]{ieeeconf}  
\pdfoutput=1
\IEEEoverridecommandlockouts                              

\usepackage[utf8]{inputenc} 
\usepackage[T1]{fontenc}    
\usepackage{hyperref}       
\usepackage{url}            
\usepackage{booktabs}       
\usepackage{amsfonts}       
\usepackage{amsmath}
\usepackage{graphicx}
\usepackage{nicefrac}       
\usepackage{microtype}      
\usepackage[table]{xcolor}         
\usepackage{subfiles} 
\usepackage{bm}
\usepackage{stmaryrd}
\usepackage{wrapfig}
\usepackage{comment}
\usepackage{balance}
\usepackage{tablefootnote}
\usepackage[linesnumbered,lined,ruled]{algorithm2e}
\usepackage{pifont}
\newcommand{\cmark}{\ding{51}}%

\SetCommentSty{mycommfont}

\title{\bf \LARGE Enforcing the consensus between Trajectory Optimization  \\ and Policy Learning for precise robot control}

\newcommand{\subjto}{\mathrm{s.t.}}
\newcommand{\almsure}{\mathrm{a.s.}}

\newcommand{\bmu}{\boldsymbol{u}}
\newcommand{\bmx}{\boldsymbol{x}}
\newcommand{\bmlambda}{\boldsymbol{\lambda}}
\newcommand{\calA}{\mathcal{A}}
\newcommand{\calP}{\mathcal{P}}
\newcommand{\calL}{\mathcal{L}}
\newcommand{\calX}{\mathcal{X}}

\newcommand{\bbEE}{\mathbb{E}}

\DeclareMathOperator*{\argmin}{arg\,min}

\author{%
  Quentin Le Lidec\textsuperscript{*,1} \and Wilson Jallet\textsuperscript{1,2}\and Ivan Laptev\textsuperscript{1} \and Cordelia Schmid\textsuperscript{1} \and Justin Carpentier\textsuperscript{1}%
  \thanks{%
        \textsuperscript{*}Corresponding author}
  \thanks{%
        \textsuperscript{1}Inria - Département d'Informatique de l'\'Ecole normale supérieure, PSL Research University. Email: \texttt{firstname.lastname@inria.fr}}
    \thanks{%
        \textsuperscript{2}LAAS-CNRS, 7 av.~du Colonel Roche, 31400 Toulouse}
}

\begin{document}

\maketitle

\begin{abstract}
  Reinforcement learning (RL) and trajectory optimization (TO) present strong complementary advantages.
  On one hand, RL approaches are able to learn global control policies directly from data, but generally require large sample sizes to properly converge towards feasible policies.
  On the other hand, TO methods are able to exploit gradient-based information extracted from simulators to quickly converge towards a locally optimal control trajectory which is only valid within the vicinity of the solution.
  Over the past decade, several approaches have aimed to adequately combine the two classes of methods in order to obtain the best of both worlds.
  Following on from this line of research,  we propose several improvements on top of these approaches to learn global control policies quicker, notably by leveraging sensitivity information stemming from TO methods via Sobolev learning, and Augmented Lagrangian (AL) techniques to enforce the consensus between TO and policy learning.
  We evaluate the benefits of these improvements on various classical tasks in robotics through comparison with existing approaches in the literature.
\end{abstract}

\section{Introduction}\label{sec:intro}

\PARstart{B}{y} leveraging derivative information from the dynamics and costs, optimal control (OC) algorithms~\cite{li2004iterative,mayneSecondorderGradientMethod1966,tassaSynthesisStabilizationComplex2012,mastalli2020crocoddyl} efficiently compute local controllers.
Model Predictive Control (MPC)~\cite{rawlings2017model} aims at retrieving a local controller with state feedback through online re-planning.
However, current algorithms remain computationally too expensive to be run at high frequencies, are sensible to local minima and integration of information from sensors (e.g. force, vision) is technically difficult.
This limits both their real-world capabilities and the practicality of deployment on complex settings requiring various sensor modalities.
For these reasons, control policies learned directly from data are appealing as they offer the possibility to perform sensor-fusion and executed at very high frequencies.

Reinforcement learning (RL)~\cite{sutton2018reinforcement} is the prevailing paradigm when it comes to learning policies directly from observations.
In RL, the physical dynamics are usually considered as unknown and classical algorithms~\cite{williams1992simple} rely on stochastic policies to build $0$\textsuperscript{th}-order gradient estimators.
Although they are unbiased, the high variance of such estimators make them very sample-inefficient.
Actor critic approaches~\cite{sutton1999policy} combine ideas from policy gradient and approximate dynamic programming by using an estimate of the Q function to reduce variance of policy gradient algorithms.
This improves the sample efficiency but often at the cost of less stable training, and more advanced strategies, \textit{e.g.} Proximal Policy Optimization (PPO)~\cite{schulman2017proximal} or Soft Actor-Critic (SAC)~\cite{haarnoja2018soft}, are often focused on regaining more stability.
Eventually, even on problems easily solved via standard \textit{model-based} approaches such as LQR, model-free algorithms exhibit several limitations and simpler rather than more complex methods may perform better~\cite{mania2018simple}.

In robotics, RL is never truly model-free, as samples are often generated by simulators which are actually based on models of physics.
Therefore, we propose to embrace models when they are available, and, by using gradient information, to move from $0$\textsuperscript{th}-order to $1$\textsuperscript{st}-order methods with superlinear convergence properties.
Such an approach is made possible by the recent emergence of differentiable physics engines, relying on derivatives of rigid body algorithms~\cite{carpentier2018analytical} and frictional contacts \cite{de2018end,lelidec2021differentiable}.
A second key ingredient for model-based approaches is access to efficient numerical solvers for trajectory optimization~\cite{mastalli2020crocoddyl,jalletImplicitDifferentialDynamic2022}.
Indeed, specialized OC algorithms exploit physical models, their derivatives, and the inherent temporal structure of the problem via the Bellman equation (or Pontryagin's principle).
In particular, quasi-Newton (\textit{e.g.,} iterative Linear Quadratic Regulator (iLQR)~\cite{li2004iterative,tassaSynthesisStabilizationComplex2012}) or second-order methods (\textit{e.g.}, Differential Dynamic Programming (DDP)~\cite{mayneSecondorderGradientMethod1966}), lead to improved convergence rates.

For complex scenarios, OC is often too slow to be run online, especially when it includes constraints on the system.
Yet, it remains attractive to draw benefits from its capabilities offline; typically, for training policies.
Following this path, guided policy search (GPS)~\cite{levine2013guided, levine2016end} aims at combining the strengths of OC and policy learning.
Closer to our work, \cite{mordatch2014combining} considers known deterministic dynamics and uses an ADMM framework to embed the policy learning problem within the OC formulation.
The dual update from ADMM allows to automatically and gradually reach a consensus between the learned policy and the local controllers that OC yields.
However, they relax the agreement constraint between RL and OC, which results in shifted, thus degraded, solutions.
Additionally, they further exploit the feedback gains from trajectory optimization in order to supervise the first order derivatives of the policy.
This technique, also known as Sobolev training~\cite{simard1998transformation, czarnecki2017sobolev}, regularizes the training and improves generalization and thus the stability of the control policy.
In this work, we extend it by using a stochastic variant~\cite{czarnecki2017sobolev} to boost computational efficiency.
Finally, a more recent work~\cite{mora2021pods} improves rollouts of the policy by performing approximate Newton steps in the trajectory space.
Here we establish some links between such an approach and OC, which in turn allows us to set the stage for a general formulation incorporating constraints while resulting in more efficient algorithms.
Table~\ref{tab:methods_char} summarizes some characteristics of our approach and the related works, \textit{i.e.} the precision of the learned policy (Precision),  the number of samples required to converge (Efficient), the ability to handle physical constraints (Constraints) or to leverage higher order information (Sobolev) and the need for a physical model (Model-free).

In this paper, we first present a general formulation, including physical and geometrical constraints, for policy learning which lays the ground for a better interplay between OC and policy learning (Sec.~\ref{sec:formulation}).
In Sec.~\ref{sec:pddp}, we propose a first algorithm based on a projected descent and detail how stochastic Sobolev learning can be leveraged to provide a computationally efficient higher-order supervision of policy training.
A second variation based on Augmented Lagrangian (AL) techniques is introduced in Sec.~\ref{sec:al_learning} to automatically and precisely reach a consensus between the policy and the trajectory optimization oracle.
To properly account for the lack of consensus at the beginning of the learning process, we propose to add multiple shooting component which allows to sensibly reduce the computational timings of learning a control policy.
Finally, in Sec.~\ref{sec:exp}, the influence of each component of the approach is evaluated separately and we validate the final algorithms on a set of standard robotics tasks.

\begin{table}
  \caption{Characteristics of various policy learning algorithms.}
  \label{tab:methods_char}
  \centering
  \resizebox{\columnwidth}{!}{
    \begin{tabular}{l|ccccc}
      \toprule
                                       & Precise  & Efficient & Constraints & Sobolev & Model-free  \\
      \midrule
      RL                            &  &     &       &   & \cmark     \\
      \rowcolor{lightgray}
      DPL \cite{mordatch2014combining} &  & \cmark    &       & \cmark  &      \\
      PODS \cite{mora2021pods}         & \cmark & \cmark  &       &   &     \\
      \rowcolor{lightgray}
      AL (Ours)                        & \cmark & \cmark    & \cmark      & \cmark  &      \\
      PDDP (Ours)                       & \cmark & \cmark    & \cmark      & \cmark  &      \\
      \bottomrule
    \end{tabular}
  }
  \vspace{-0.5cm}
\end{table}

\section{Interplay between policy learning and trajectory optimization} \label{sec:formulation}

We consider deterministic and known system dynamics
\begin{subequations}
\label{eq:ocp_cons}
\begin{align}
    \label{eq:traj_cons}
    & x_{t+1} = f(x_t, u_t)\\
    &
    \label{eq:state_control_constraints}
    x_t \in \calX \text{ and } u_t \in \mathcal{U} 
\end{align}
\end{subequations}
while other parameters of the optimal control problem (OCP) are randomly distributed. 
These parameters, denoted by $\beta$, include the initial condition of the problem $x^0 \in\calX$, as well as other parameters (desired terminal state or end-effector position); they follow a random distribution $\calP$.
The set $\mathcal{U} = [u_{\min},u_{\max}]$ contains the admissible control inputs, of dimension $n_u$, while $\mathcal{X}$ refers to the set of feasible states also accounting for obstacles.
The deterministic assumption on the dynamics is reasonable, as a wide range of systems (robotic arms, manipulators, quadrupeds, drones) can accurately be described without stochasticity.
We denote \mbox{$\bmu = (u_0, \ldots, u_{T-1})$} the sequence of controls, and \mbox{$\bmx = (x_0,\ldots,x_T)$} the state sequence.
The total cost associated with a given control sequence $\bmu$ is:
\begin{equation}
\label{eq:trajectory_cost}
        R(\beta; \bmu)     = \sum_{t=0}^{T-1} ~ \ell_t(x_t,u_t; \beta) + \ell_T(x_T; \beta)
\end{equation}
Unlike previous works in the literature \cite{mordatch2014combining, mora2021pods}, we also account in this work for state and control constraints in Eq.~ \eqref{eq:state_control_constraints} directly in the TO problems.
Experiments from Sec.~\ref{sec:exp} notably demonstrate how this consideration of the constraints substantially improves the precision and the stability of the resulting policy.

The goal of this paper is to learn closed-loop control policies $\pi\colon  \calX \to \mathcal{U}$ which approximately minimizes the trajectory cost.
The policy $\pi$ is constrained  to lie in a set of parameterized policies $\Pi$; typically, we will consider neural network policies with two hidden layers of 256 units each.
The policy parameters are denoted by $\theta$ and lie in a set $\Theta$.
The policy space is thus
\[
    \Pi = \{ \pi_\theta: \theta \in \Theta \}.
\]
The activation functions are ReLU except for the final layer which is activated by a hyperbolic tangent so that the output is restricted to $ [ u_{\min}, u_{\max} ]$.
We denote by $R$, the cost function and $J(\beta;\pi_\theta)$ the \textit{policy cost}, which is the trajectory cost with controls given by $u_t = \pi_\theta(x_t)$:
\begin{equation}
\begin{aligned}
    &J(\beta; \pi_\theta) = R\left(\beta; \pi_\theta(x_0), \ldots, \pi_\theta(x_{T-1})\right) \\
    &\text{with } x_0 = x^0, \ 
    x_{t+1} = f(x_t, \pi_\theta(x_t))
\end{aligned}
\end{equation}
Our goal is to solve the \emph{policy optimization problem}:
\begin{equation}
    \label{eq:policy_optimization_1}
    \min_{\theta\in\Theta} \bbEE_\beta\left[ J(\beta; \pi_\theta)) \right]
\end{equation}
where $\bbEE_\beta[\cdot]$ is the expectation. 

To do so, we will follow the path initiated in \cite{mordatch2014combining, levine2014learning} and adopt a constrained formulation linking together trajectory optimization and policy optimization:
\begin{subequations}
    \label{eq:constrained_formulation}
    \begin{align}\label{eq:constr_form_a}
        \min_{\theta, \bmu} \  & \bbEE_\beta\left[ R(\beta; \bmu) \right]                                                                         \\
        \subjto  \             & \forall t \ u_t = \pi_\theta(x_t), \ \almsure
    \end{align}
\end{subequations}
where we now optimize over a \textit{distribution} of control sequences $\bmu$, and $\bmx = (x_t)_t$ is defined by \eqref{eq:traj_cons}. The constraint enforces that the controls and the policy are consistent on almost all states $x_t$ reached by randomization of $\beta$.

\section{Alternating between OC and Supervised learning: a projected DDP approach} \label{sec:pddp}

\noindent
\textbf{Projected DDP algorithm.} A classical way of solving a constrained optimization problem consists in alternating between making a step on the unconstrained problem and projecting the new iterate on the constraints (e.g., projected gradient algorithms).
In this spirit, \eqref{eq:constrained_formulation} can be solved by optimizing the control $\bmu$ via trajectory optimization before projecting the new trajectory on the set of trajectories obtainable through $\Pi$.
Discarding the constraints, the problem \eqref{eq:constrained_formulation} can be solved by approximating the expectation with a finite sum $\hat{R}(\bmu)$ obtained by sampling $N$ instances of the problem $\beta^{(i)} \sim\calP$:
\begin{equation}
    \hat{R}(\bm{U}) = \frac{1}{N}\sum_{i=1}^{N} R(\beta^{(i)}; \bmu^{(i)}).
\end{equation}
where $\bm{U} = (\bmu^{(1)}, \ldots, \bmu^{(N)})$ is the entire set of control sequences. The objective above is separable, hence each $\bmu^{(i)}$ can be optimized separately in parallel.

The sampled problems are then optimized via discrete-time trajectory optimization methods such as Differential Dynamic Programming (DDP)~\cite{mayneSecondorderGradientMethod1966} or its variant iLQR~\cite{tassaSynthesisStabilizationComplex2012}, using a rollout originating from the learned policy $\pi_\theta$ as a warm-start.
Then, the local control trajectories $(\bmu^{(i)})_i$ are projected onto an element of $\Pi$ in the least-squares sense. The projection problem is equivalent to the following supervised learning problem:
\begin{equation}
    \min_{\theta\in\Theta} \hat{d}(\theta, \bm{U}) = \sum_{i=1}^N\sum_{t=0}^{T-1}
    \frac{1}{2}\left \| u^{(i)}_t  - \pi_\theta \big(x_t^{(i)}\big)\right \|^2_2.
\end{equation}
which is solved using classical stochastic mini-batch gradient algorithms such as Adam \cite{kingma2014adam}.
Alternating between these two operations results in Alg.~\ref{alg:pddp} which we call Projected DDP (PDDP).

\SetKwRepeat{Do}{do}{while}
\begin{algorithm}[t]
    \SetAlgoLined
    \KwIn{Distribution on the parameters $\beta \sim \calP$, model for the policy: $\pi_\theta \in \Pi$}
    \KwOut{Optimal policy $\pi_{\theta}^\star$}
    \For{$k=1$ \KwTo $M$}{
        $\bm U^{k+1,0} \leftarrow$ Rollout$(\pi_{\theta^k})$\tcp*{Initial guess}
        $\bm{U}^{k+1} \leftarrow \argmin_{\bm{U}} \hat{R}(\bm{U})$\tcp*{OC (with guess)} \label{line:optimal_control1}
        \tcp{Supervised learning (can also use $\hat{d}_2$)}
        $\theta^{k+1} \leftarrow \argmin_\theta \hat{d}(\theta, \bm{U}^{k+1})$\; \label{line:supervised_learning1}
    }
    \caption{Projected DDP descent (PDDP)}
    \label{alg:pddp}
\end{algorithm}

Such an approach encompasses the one proposed in \cite{mora2021pods} where Gauss-Newton (GN) steps are successively projected.
By establishing an interplay with OC, here we rather perform a DDP step which is just a more efficient way to compute a GN step that also handles the constraints from \eqref{eq:ocp_cons}.
Previous works \cite{mordatch2014combining, mora2021pods} often ignore these constraints during the trajectory optimization phase.
Typically, the box constraint on the control is rather lifted to the policy space via a tanh activation function on the last layer.
This discrepancy can be fatal as trajectories obtained from OC might not be reproducible by the neural network and, thus, could cause divergence of the combined approach.
Experiments in Sec.~\ref{sec:exp} demonstrate how crucial this specificity of our methods is to ensure convergence.
Another obvious advantage is the possibility to consider more complex tasks involving geometric constraints or dynamic constraints on the systems to control (\textit{e.g.}, obstacles or joint limits).

\vspace{0.2cm}
\noindent
\textbf{Stochastic Sobolev learning.}
If the constraint from \eqref{eq:constrained_formulation} enforces the output of the learned policy to match the local controllers, this should also be true for higher order derivatives.
This idea, exploited in \cite{mordatch2014combining}, refers to the concept of Sobolev training~\cite{czarnecki2017sobolev}.
Second-order methods such as DDP additionally yield local \textit{feedback} gains $K_t$ around the optimized trajectory. 
These gains can be used for stabilization of the controlled system around the local optimum, as~\cite{dantecFirstOrderApproximation2022} has shown on real-world systems.
For policy learning, we can make use of these gains to regularize the policy $\pi_\theta$ during the supervised learning phase (Alg.~\ref{alg:pddp}, line~\ref{line:supervised_learning1}), leading to the following supervised learning problem:
\begin{equation}
    \label{eq:sobolev_loss}
    \min_{\theta\in\Theta} \hat{d}_2(\theta, \bm{U}) = \hat{d}(\theta, \bm{U}) + \frac{1}{2}\sum_{i,t}
    \bigg\| K^{(i)}_t  - \frac{\partial \pi_\theta}{\partial x} (x_t^{(i)}) \bigg\|^2_2
\end{equation}

However, getting the second order derivative $\partial_{x\theta}\pi_\theta$, as in \cite{mordatch2014combining}, is computationally expensive as it requires to perform multiple backpropagation operations.
For this reason, we propose to use a stochastic version of Sobolev learning \cite{czarnecki2017sobolev} which is done by matching projections of $K$ and $\partial_x\pi_\theta$ on control directions $v^{(i)}$ randomly sampled on the unit sphere. The regularization term from \eqref{eq:sobolev_loss} becomes $\frac{1}{2} \sum_{i,t} \| (v^{(i)})^\top K^{(i)}_t  - \partial_x \big({v^{(i)}}^\top\pi_\theta (x_t^{(i)}) \big)\|^2_2$ and this greatly improves computational efficiency by avoiding backpropagating several times across the neural network.

\section{Gradually enforcing a consensus: an Augmented Lagrangian approach}\label{sec:al_learning}

The Lagrangian function associated to the constrained formulation \eqref{eq:constrained_formulation} is given by:
\begin{equation}
    \mathcal{L}(\theta, \bmu,  \bmlambda) = \bbEE_\beta\left[ R(\beta; \bmu)  + \sum_{t=0}^{T-1} \lambda_t^\top \left( u_t - \pi_\theta(x_t) \right) \right]
\end{equation}
where $\bmlambda = (\lambda_t)_{0\leq t\leq T} $ now designate random variables which are the Lagrange multipliers.
We can define the corresponding augmented Lagrangian by:
\begin{equation}
        \label{eq:augmented_lagrangian}
        \calL_A^\mu(\theta, \bmu,  \bmlambda) = \calL(\theta,\bmu,\bmlambda) + \bbEE_\beta\left[ \sum_{t=0}^{T-1} \frac{\mu_t}{2}\left\| u_t - \pi_\theta(x_t) \right\|^2 \right]
\end{equation}

\vspace{0.2cm}
\noindent
\textbf{ADMM algorithm.} 
We first consider that a set of sample parameters $(\beta^{(i)})$ is fixed during training. 
Indeed, the Lagrange multipliers are intrinsically linked to the sampled values of $\beta$ and thus, re-sampling them would require to update $\bmlambda$ without losing progress on the optimization, which is a difficult task left as future work.
This differs from Alg.~\ref{alg:pddp} which can be naturally run online.
Thus, we consider the following sampled version of the constrained problem~\eqref{eq:constrained_formulation}:
\begin{subequations}
    \label{eq:sampled_constrained_gps}
\begin{align}
    \min_{\theta, \bm{U}} \  & \frac{1}{N}\sum_{i=1}^N R(\beta^{(i)};\bmu^{(i)}) \\
    \label{eq:sampled_constrained_gps:b}
    \subjto  \             & u_t^{(i)} = \pi_\theta(x_t^{(i)}), \ \forall t,\ \forall i \in \llbracket 1, N\rrbracket
\end{align}
\end{subequations}
where $(\beta^{(i)})_{1\leq i\leq N}$. 
The classical and augmented Lagrangians for this sampled variant are respectivily given by:
\begin{align}
    &\hat{\calL}(\theta, \bm U, \bm\Lambda) = \hat{R}(\bm U) + \sum_{i,t} (\lambda_t^{(i)})^\top (u_t^{(i)}-\pi_\theta(x_t^{(i)})), \\
    \label{eq:policy_augmented_lagrangian}
    &\hat{\calL}_A^\mu(\theta, \bm{U},  \bm{\Lambda}) =
        \hat{\calL}(\theta, \bm{U},  \bm{\Lambda}) +
    \sum_{i,t} \frac{\mu_t}{2} \big\| u_t^{(i)} - \pi_\theta(x_t^{(i)}) \big\|^2,
\end{align}
where $\bm{\Lambda} = (\bmlambda^{(1)}, \ldots, \bmlambda^{(N)})$ are the Lagrange multipliers for each constraint block $\{ u_t^{(i)} = \pi_\theta(x_t^{(i)})\}_t$. Both functions are separable with respect to the entry $\bmu^{(i)}$.

The problem is solved using the alternating direction method of multipliers (ADMM)~\cite{boyd2011distributed}, as described in Alg.~\ref{alg:algps} which we call PLAL. The algorithm alternatively minimizes $\calL_\calA^\mu$  w.r.t. the primal variables $\bm U$ and $\theta$, before updating the dual variable $\bm\Lambda$ through a simple dual ascent step. 
As done in Sec.~\ref{sec:pddp}, minimization w.r.t. each $\bmu^{(i)}$ is done using a DDP-type algorithm such as FDDP~\cite{mastalli2020crocoddyl}, while the minimization w.r.t. $ \theta$ is performed using a classical stochastic optimization algorithm~\cite{kingma2014adam}.
If \cite{mordatch2014combining} also uses ADMM, the consensus is approximated via a quadratic penalty term, while in this work we enforce it with a hard constraint \eqref{eq:sampled_constrained_gps:b}.
This results in more stable and precise solutions as shown in Sec.~\ref{sec:exp}.

\begin{algorithm}[t]
    \SetAlgoLined
    \KwIn{Distribution on the parameters $\beta \sim \calP$, model for the policy: $\pi_\theta \in \Pi$}
    \KwOut{Optimal policy $\pi_{\theta^\star}$}
    \For{$k=1$ \KwTo $M$}{
        $\bm U^{k+1} \leftarrow \argmin_u \hat\calL_A^\mu(\theta^k, \bm{U},  \bm{\Lambda}^k)$\; \label{line:optimal_control}
        \tcp{Supervised learning}
        $\theta^{k+1} \leftarrow \argmin_\theta \hat\calL_A^\mu(\theta, \bm{U}^{k+1},  \bm{\Lambda}^k)$\; \label{line:supervised_learning}
        \tcp{Dual update}
        $\bm\Lambda^{k+1} \leftarrow (\lambda_t^{(i)} + \mu(u_t^{(i)} - \pi_\theta(x_t^{(i)})))_{t,i}$\; \label{line:dual_update}
    }
    \caption{Policy learning via ADMM (PLAL)}
    \label{alg:algps}
\end{algorithm}

Contrary to Alg.~\ref{alg:pddp}, the OC (Alg.~\ref{alg:algps}, line~\ref{line:optimal_control}) and supervised learning (Alg.~\ref{alg:algps}, line~\ref{line:supervised_learning}) phases are now linked via the Lagrange multipliers $\bm\Lambda$.
Indeed, the additional terms appearing in \eqref{eq:policy_augmented_lagrangian} enforces the demonstrations from OC to be adapted to the current capabilities of the policy, while the dual update (Alg.~\ref{alg:algps}, line~\ref{line:dual_update}) gradually leads to an agreement between the local controllers from OC and the policy.
The previously introduced stochastic Sobolev term can naturally be integrated in the approach as a regularization of the learning phase (Alg.~\ref{alg:algps}, line~\ref{line:supervised_learning}), and the algorithm is then interpretable as an occurrence of \textit{Global Variable Consensus with Regularization}~\cite{boyd2011distributed}.

\vspace{0.2cm}
\noindent
\textbf{Multiple shooting formulation.} 
Due to the augmented Lagrangian terms, the trajectory optimization phase (Alg.~\ref{alg:algps}, line~\ref{line:optimal_control}) requires to optimize the state $\bmx$ through the neural network policy and thus involves the costly computation of the Jacobian $\partial_x\pi_\theta$.
Inspired by~\cite{mordatch2014combining}, we propose to decouple the state variables of the supervised learning and OC problems, in a way similar to what is done in multiple shooting~\cite{bock1984multiple}. 
To do so, two state variables are introduced (${^{1}}\bm X = ({^{1}}\bmx^{(i)})_i$ and ${^{2}}\bm X = ({^{2}}\bmx^{(i)})_i$):
\begin{subequations}
    \label{eq:ms_gps}
    \begin{align}
        \min_{\theta, \bm U, {^1}\bm X, {^2}\bm X} \  & \frac{1}{N}\sum_{i=1}^N R(\beta^{(i)}; {^1}\bmx^{(i)}, \bmu^{(i)})                 \\
        \subjto  \             & \forall t,\forall i, \ u_t^{(i)} = \pi_\theta({^2}x_t^{(i)}) \\
                            & {^2}\bm X = {^1}\bm X
    \end{align}
\end{subequations}
The corresponding Augmented Lagrangian is:
\begin{equation}
        \begin{split}
            \calL_A^\mu&{}(\theta, {^1}\bm X, {^2}\bm X, \bm U,  \bm\Lambda, \bm\Gamma)
            = \frac{1}{N}\sum_{i=1}^N\Bigg[ R(\beta^{(i)}; {^1}\bmx^{(i)},\bmu^{(i)})  \\
            &+ \sum_{t=0}^{T-1}  \frac{\mu_t^{(i)}}{2}\left\| u_t^{(i)} + \frac{\lambda_t^{(i)}}{\mu_t^{(i)}}- \pi_\theta({^2}x_t^{(i)}) \right\|^2  \\  & +\frac{\mu_t^{(i)}}{2}\left\| {^1}x^{(i)}_t + \frac{\gamma_t^{(i)}}{\mu_t^{(i)}}- {^2}x^{(i)}_t \right\|^2 \\
            &- \frac{1}{2\mu_t^{(i)}}\left\| \lambda_t^{(i)}\right\|^2
            - \frac{1}{2\mu_t^{(i)}}\left\| \gamma_t^{(i)}\right\|^2\Bigg]
    \end{split}
\end{equation}
and the update rule for $\gamma$ is:
\begin{align}
    \label{eq:gamma_update}
    {\gamma_t^{(i)}} \leftarrow {\gamma_t^{(i)}} +  \mu_t^{(i)}( {^1}x_t^{(i)} - {^2}x^{(i)}_t )
\end{align}
By decoupling the state variables between the control and learning problems, we avoid the need of computing the Jacobian of $\pi_\theta$ when doing optimal control, lifting the burden to a simple dual update \eqref{eq:gamma_update}.

\section{Experiments}\label{sec:exp}
Our implementation uses Pinocchio~\cite{carpentier2019pinocchio} and Crocoddyl~\cite{mastalli2020crocoddyl} for defining and solving OCPs, and relies on PyTorch~\cite{paszke2017automatic} for learning the neural network.
We also propose our own implementations of PODS~\cite{mora2021pods} and DPL~\cite{mordatch2014combining} as they were not provided along with the original papers.
For RL algorithms, \textit{i.e.} PPO \cite{schulman2017proximal} and SAC \cite{haarnoja2018soft}, we use implementations provided by \cite{raffin2021stable}.
Our code is open-source and will be publicly released upon publication acceptance. Every experiment was run on a single laptop without GPUs and took about 5 minutes to train policies, even on complex systems such as UR5.

\begin{figure*}[t]
    \centering
    \setlength{\tabcolsep}{0.1mm}
    \def\mywidth{0.166\linewidth-0.1mm}

    \newcommand{\myfig}[1]{\includegraphics[width=\mywidth,trim=10em 16em 20em 16em,clip=true]{#1}}
    \begin{tabular}{cccccc}
        \myfig{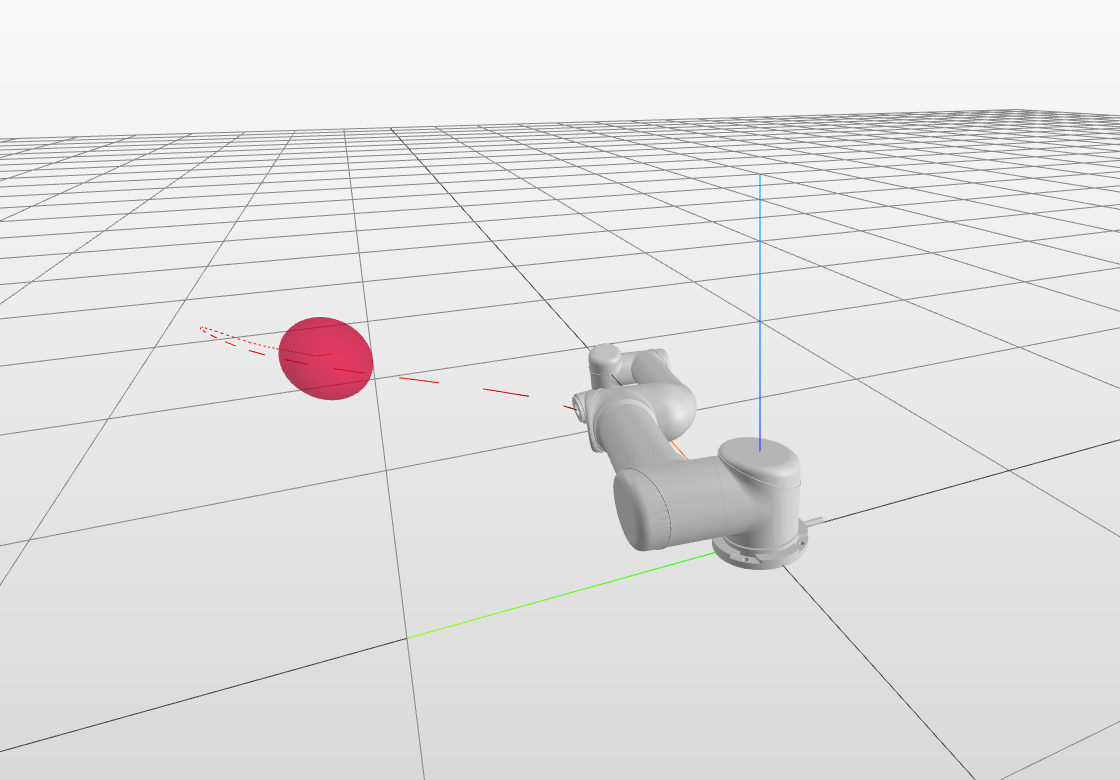}
        &
        \myfig{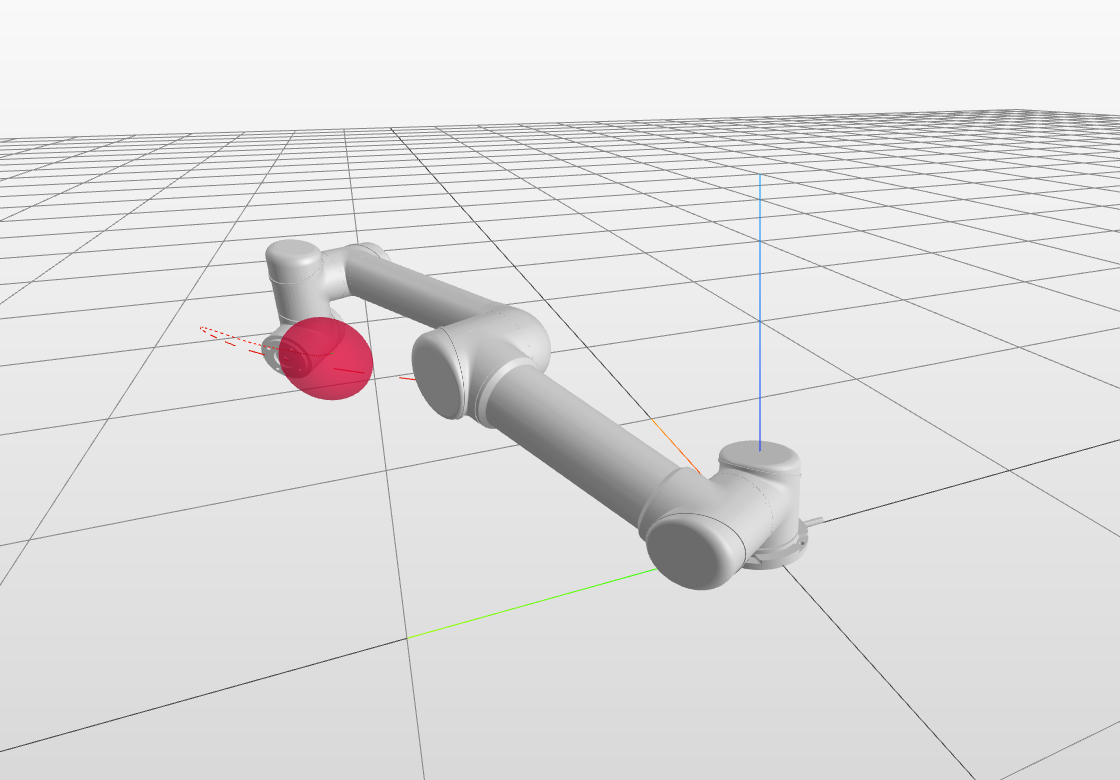}
        &
        \myfig{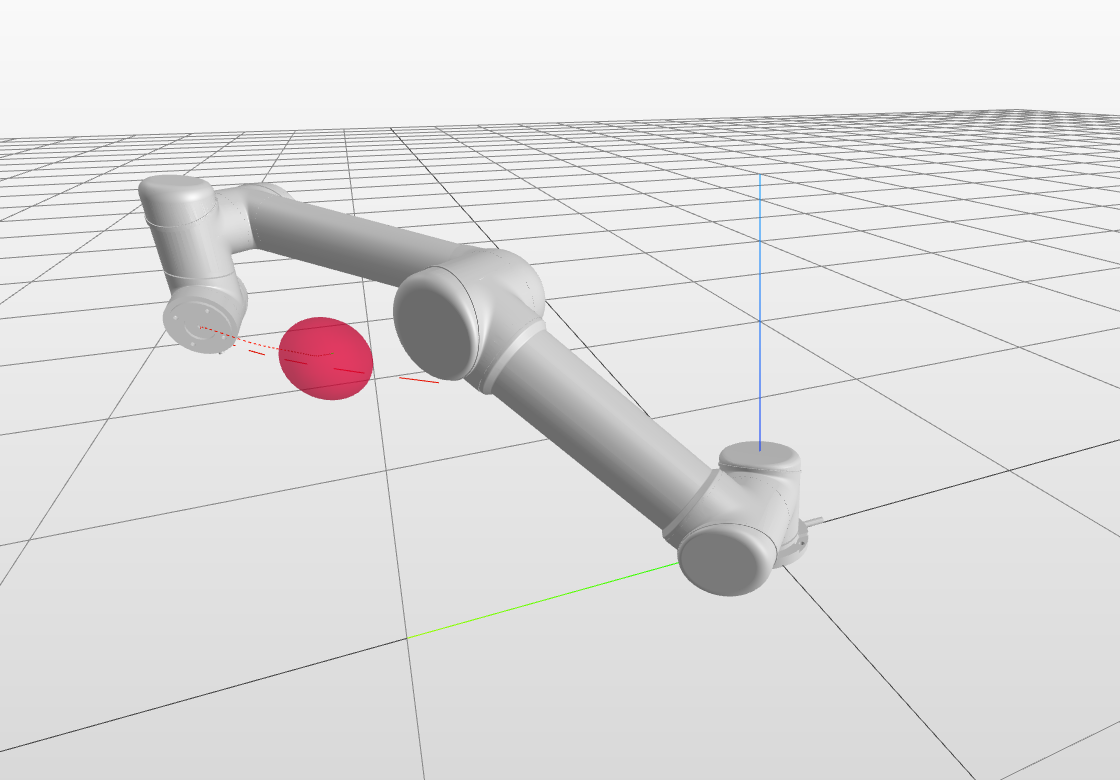}
        &
        \myfig{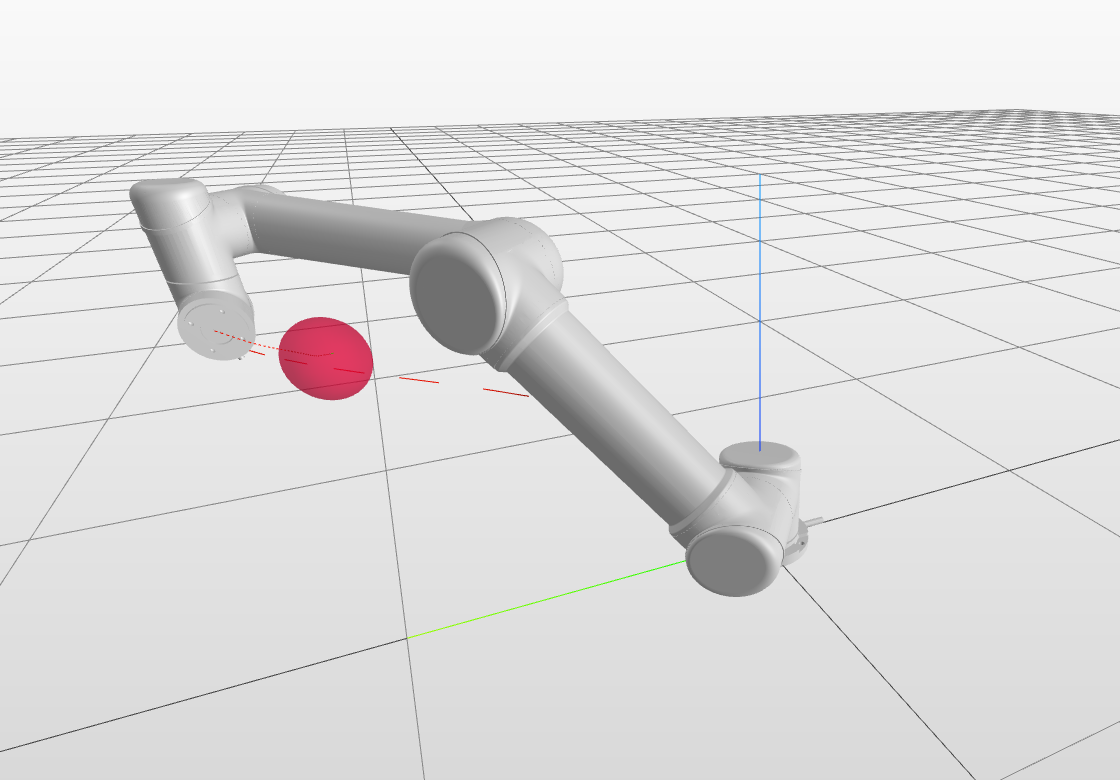}
        &
        \myfig{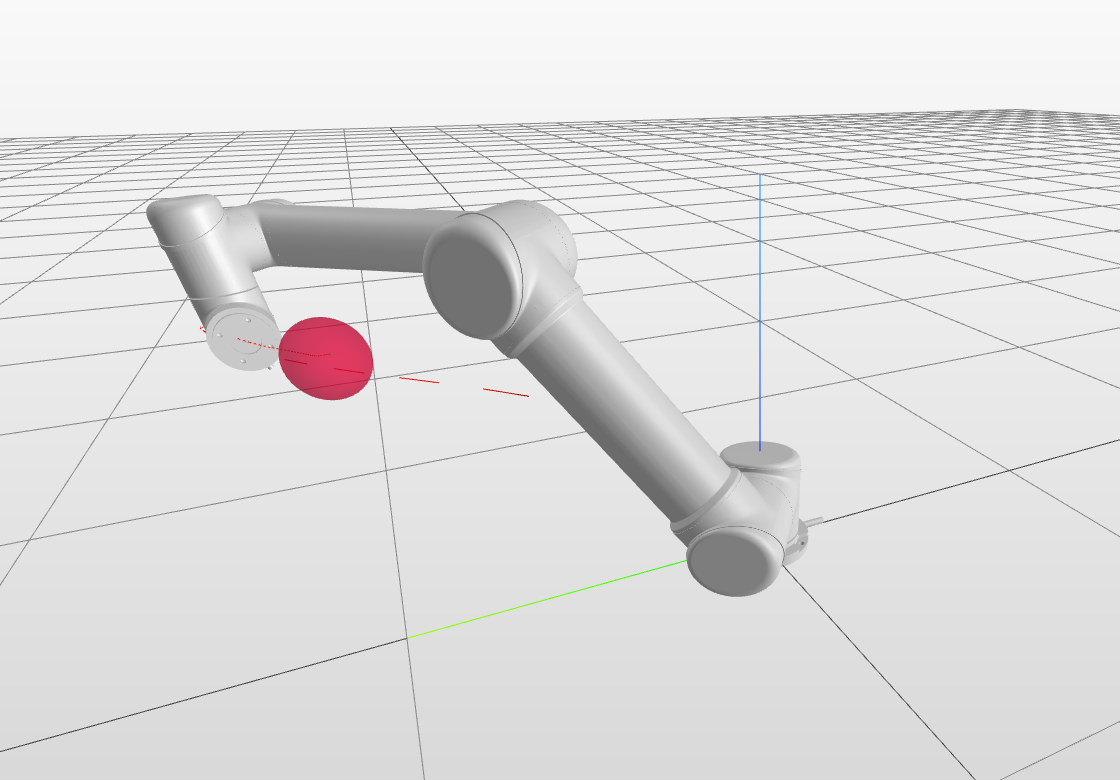}
        &
        \myfig{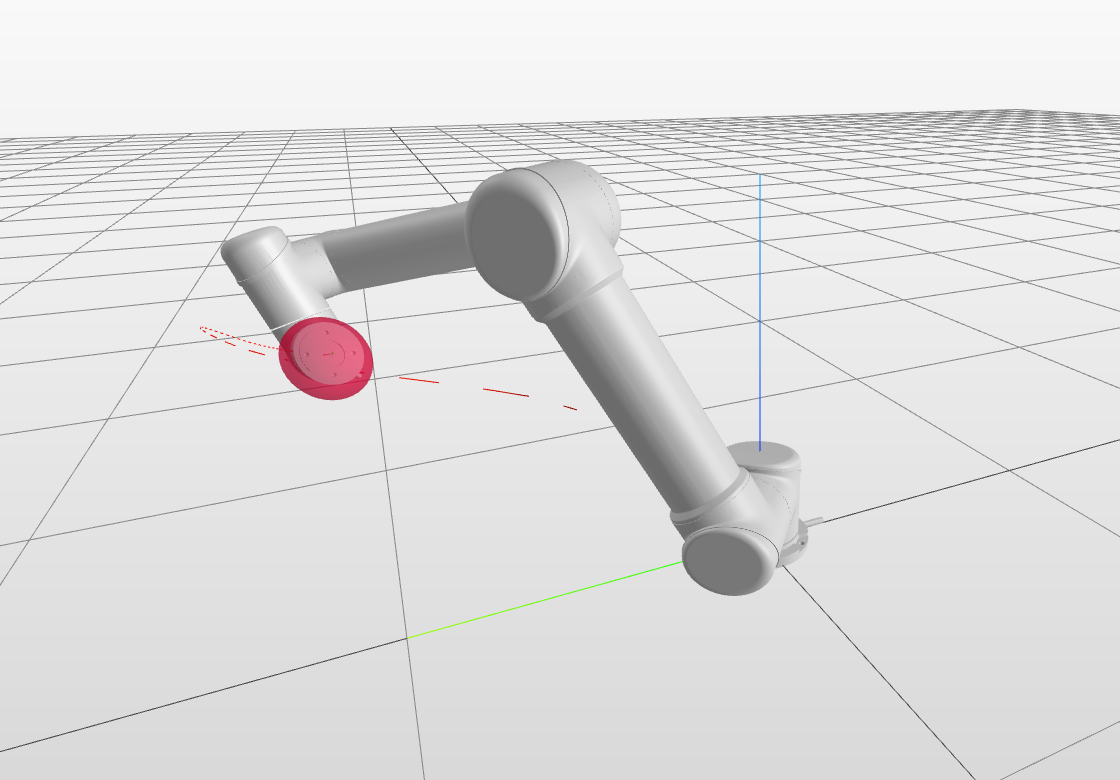}
    \end{tabular}
    \caption{\textbf{Learned policy on UR5.} After training, a rollout of the policy leads to precise control on a test problem. }\label{fig:UR5_bench}
    \vspace{-0.55cm}
\end{figure*}

The first problem we consider is a constrained LQR where the control inputs are forced to stay into a box. This is a typical convex optimization problem already very well-studied in the control community.
This control problem already highlights different characteristics and limitations of approaches under consideration.
The second set of experiments are related to robotics systems of increasing complexity: an inverted simple pendulum, a double pendulum and an UR5 robotic arm.
These problems are challenging as their dynamics are highly non-linear and the considered targets are unstable.
In addition, we consider non-linear cost functions of the form:
\begin{equation}
    \ell(x_t, u_t) = \frac{1}{2}\left  \| p(x_t) - \overline{p} \right \|_{W_p}^2 + \frac{1}{2}\left \|u_t \right \|_{W_u}^2 + \frac{1}{2} \left \| x_t \right \|_{W_x}^2 \nonumber
\end{equation}
where $p$ corresponds to the position of the end-effector which should reach a desired position $\overline{p}$ and $W_x,W_u,W_p$ are given weight matrices.
On robot systems, we only penalize the joint velocities in the state penalty term $\| x_t \|_{W_x}^2$. 

\subsection{Ablation study: constrained LQR and inverted pendulum} \label{sec:ablation}

\begin{figure}
    \centering
    \includegraphics[width=0.49\linewidth]{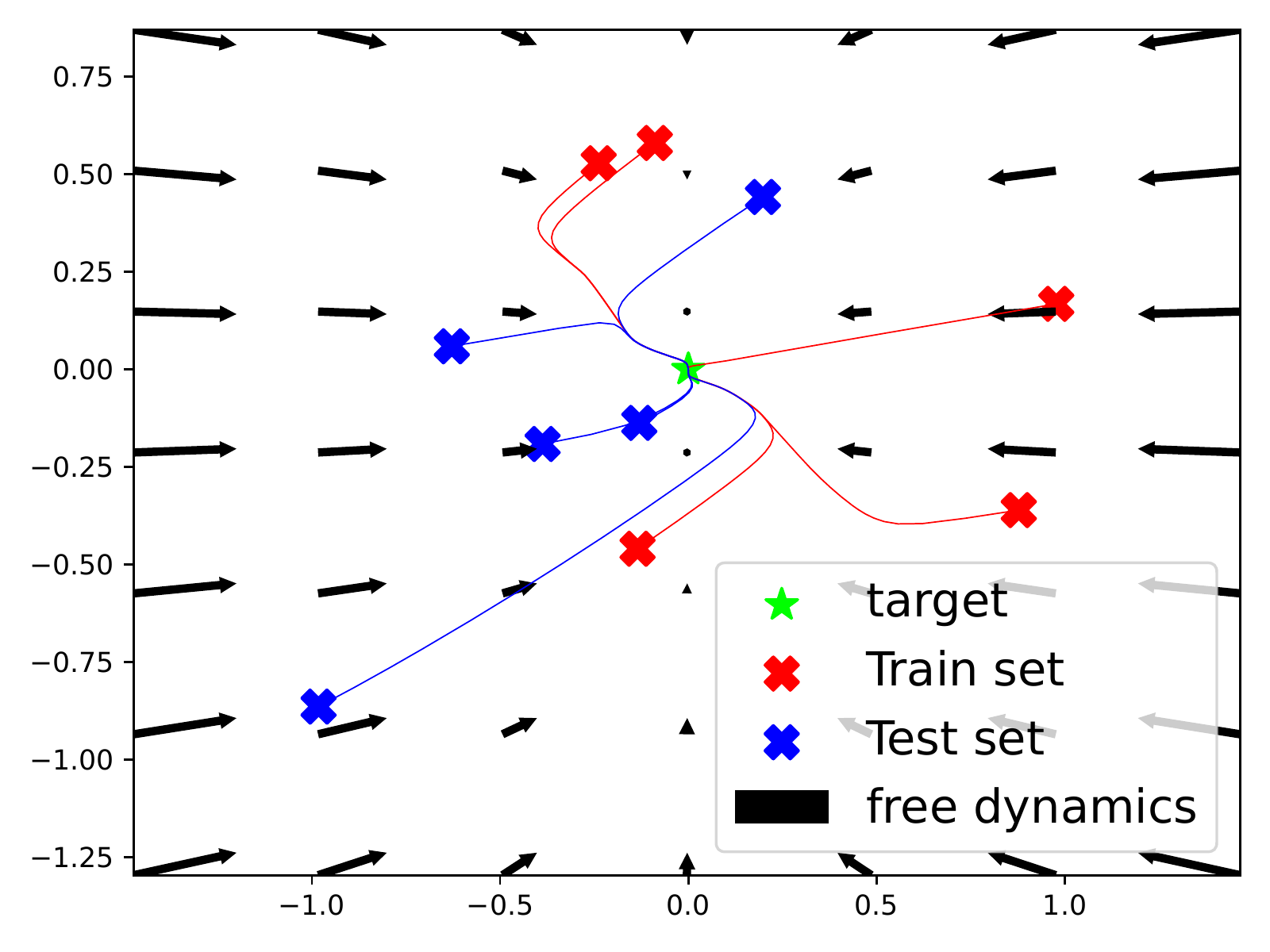}~%
    \includegraphics[width=0.49\linewidth]{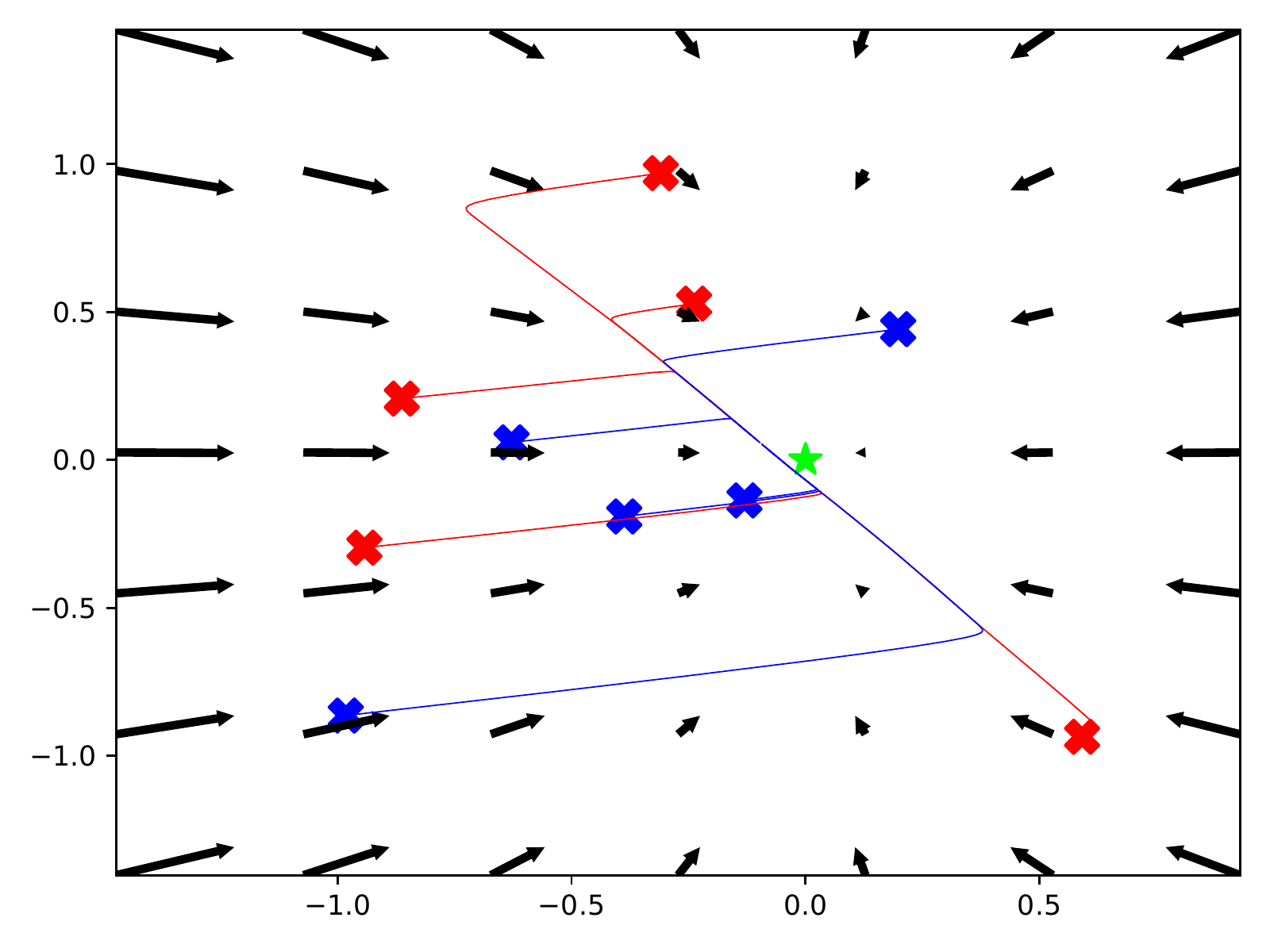}
    \caption{\textbf{Precision of the policy.} Model-based Alg.~\ref{alg:pddp}\,(\textbf{left}) leads to policies more precise than RL algorithms such as PPO\,(\textbf{right}). 
    }
    \label{fig:rl_vs_models}
    \vspace{-0.3cm}
\end{figure}

\begin{figure}[ht!]
    \centering
    \includegraphics[width=0.59\linewidth]{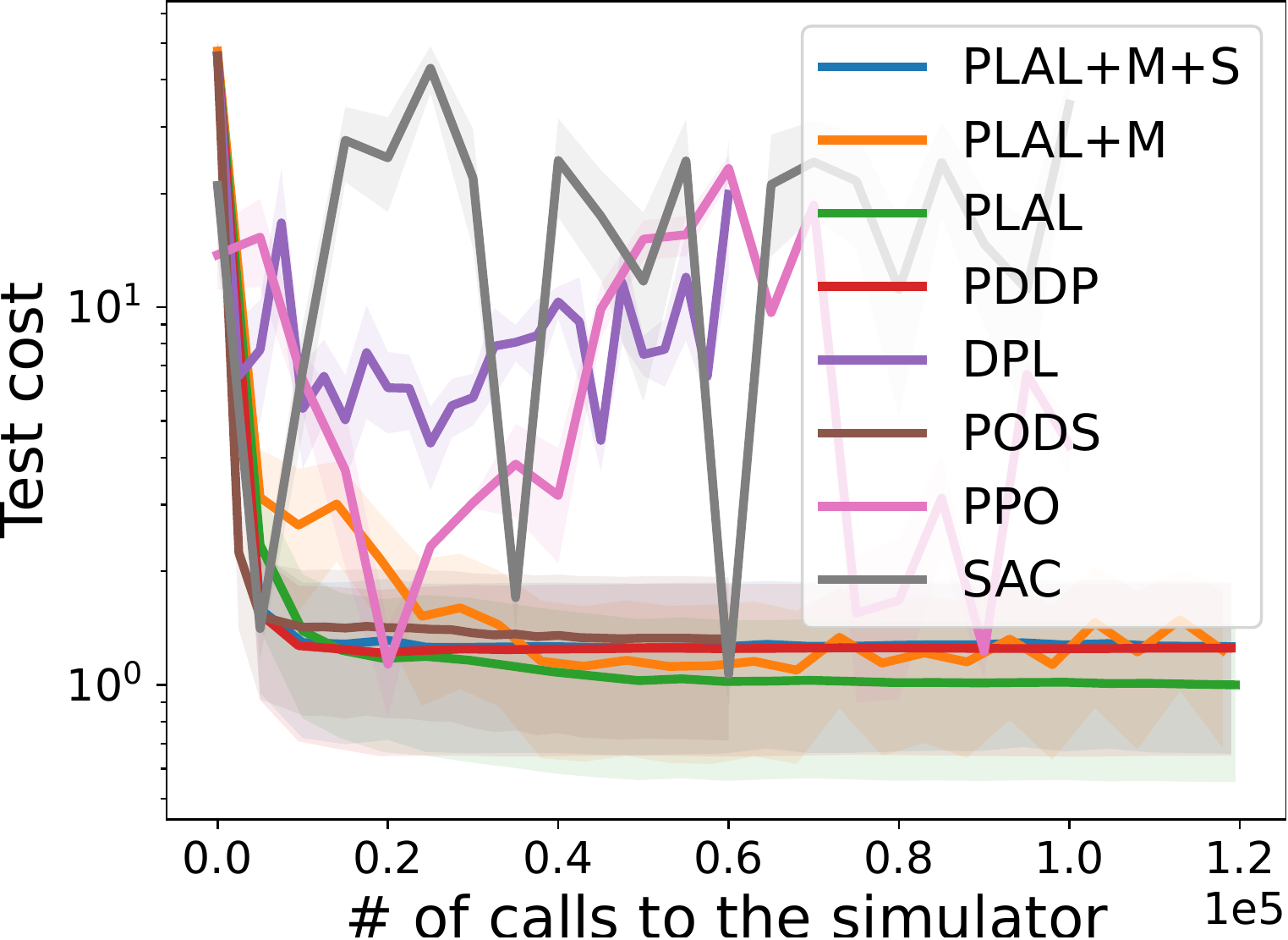}
    \caption{\small\textbf{Sample efficiency: constrained LQR.} Model-based~methods outperform model-free RL on tasks with linear dynamics and a stable target. We also evaluate the impact Sobolev training (+S) and multiple shooting (+M) or removing the constraint handling (-C).
    }
    \label{fig:LQR_bench}
    \vspace{-0.7cm}
\end{figure}

\begin{figure}[t]
    \centering
    \includegraphics[width=0.59\linewidth]{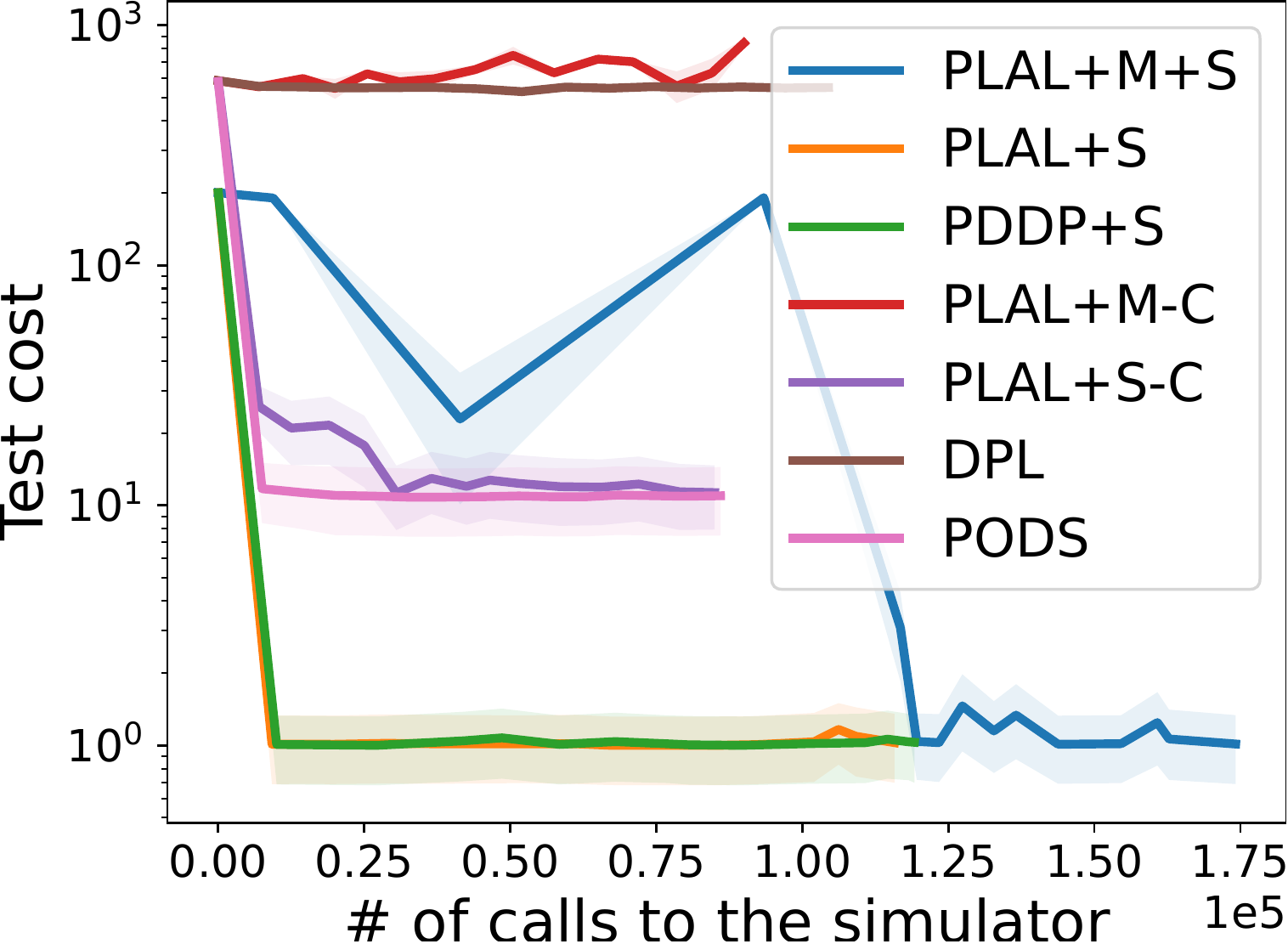}
    \caption{\small\textbf{Sample efficiency: inverted pendulum.} Using a constrained formulation \eqref{eq:constrained_formulation} along with supervision of the derivatives appears essential to reach a precise solution on a non-linear dynamics with unstable target.}
    \label{fig:pendulum_bench}
    \vspace{-0.42cm}
\end{figure}

\vspace{0.2cm}
\noindent
\textbf{Model-based vs model-free.}
Fig.~\ref{fig:LQR_bench} compares model-based algorithms, \textit{i.e.} Algs.~\ref{alg:pddp},\ref{alg:algps}, DPL and PODS, to the well-established deep RL algorithms PPO and SAC, on a constrained LQR problem.
It appears that model-based approaches are an order of magnitude more efficient than model-free algorithms.
Fig.~\ref{fig:rl_vs_models} also demonstrates how our approach is able to reach precise solutions while RL algorithms are limited by the inherent noise from $0$\textsuperscript{th}-order gradients estimates.
Among already existing model-based algorithms, PODS seems to dominate DPL both in terms of precision and efficiency.
Mainly, DPL suffers from the approximated formulation which shifts the initial problem and hinders progress towards the solution even on the constrained LQR.

\vspace{0.2cm}
\noindent
\textbf{Constrained vs unconstrained OC.} 
If including the trajectory constraints \eqref{eq:state_control_constraints} in the trajectory optimization phase allows to slightly improve results on instances of the constrained LQR (Fig.~\ref{fig:LQR_bench}), the results appear even more pronounced on the inverted pendulum problem (Fig.~\ref{fig:pendulum_bench}).
In particular, we observe a significant performance drop when removing this component (see PDDP v. PODS or PLAL+S v. PLAL+S-C on Fig.~\ref{fig:pendulum_bench}).
Intuitively, as the policy tries to mimic an unreachable control from the OC phase, it deteriorates the final solution, resulting in a significant gap in performance.

\begin{figure}[t]
    \centering
    \includegraphics[width=0.49\linewidth]{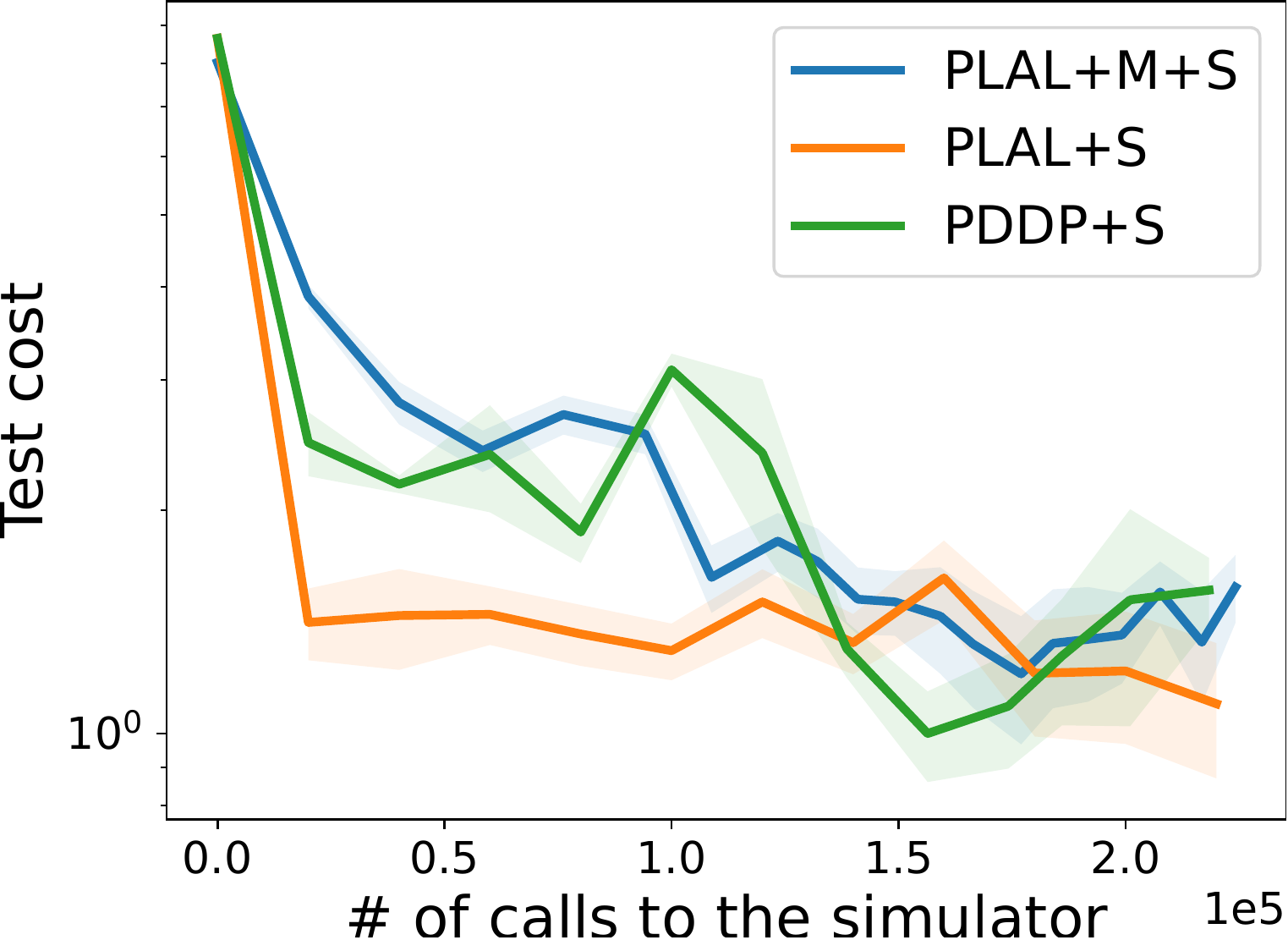}
    \includegraphics[width=0.49\linewidth]{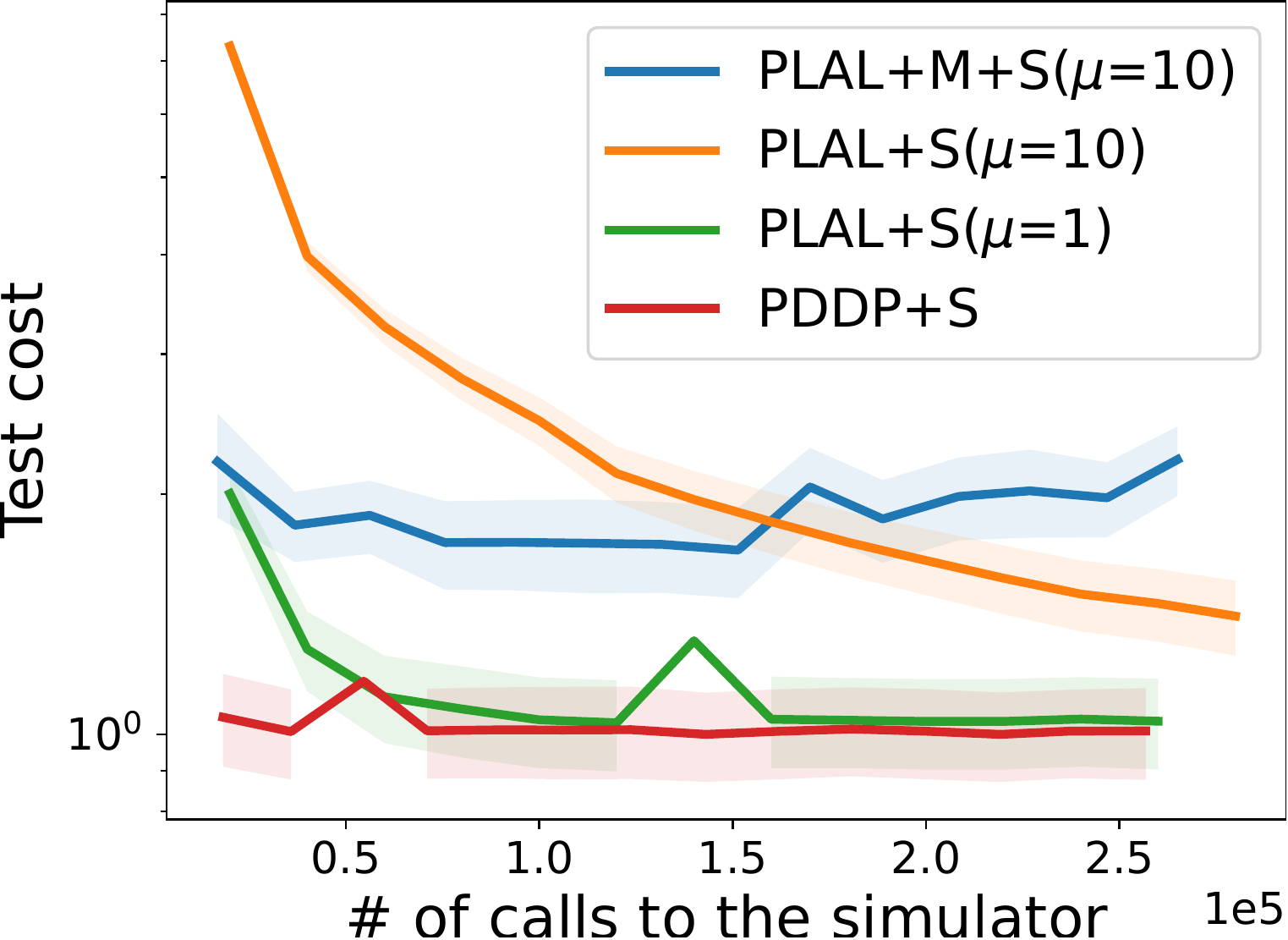}
    \caption{\small\textbf{Samples efficiency: robotics tasks.} Due to chaotic dynamics, the double pendulum (\textbf{left}) and UR5 (\textbf{right}) require both handling constraints and the Sobolev training to converge. When it is so, it converges quickly towards a precise solution. Even though policy parameters are identically initialized across the different algorithms, on UR5 the first test evaluations are only performed after a few training steps to avoid numerical divergence during rollouts.  
    }
    \label{fig:robot_bench}
    \vspace{-0.6cm}
\end{figure}

\vspace{0.2cm}
\noindent
\textbf{Multiple shooting.} 
As mentioned in Sec.~\ref{sec:al_learning}, multiple shooting allows to reduce the computational burden of the TO phase without modifying the final solution (Fig.~\ref{fig:LQR_bench}).
In practice, we observe that the OC phase of PLAL+M is 5 times faster than the one of PLAL.
However, the results from Fig.~\ref{fig:LQR_bench},\ref{fig:pendulum_bench} also exhibit how multiple shooting can require more simulator calls to learn a good policy (see PLAL+M v. PLAL on Fig.~\ref{fig:LQR_bench}, and PLAL+M+S v. PLAL+S on Fig.~\ref{fig:pendulum_bench}).

\vspace{0.2cm}
\noindent
\textbf{Sobolev training} allows to obtain gains in terms of sample efficiency by exploiting higher-order information made available by the trajectory optimization solver (see PLAL+M v. PLAL+M+S on Fig.~\ref{fig:LQR_bench}).
In addition, it also yields better generalization properties which significantly improve the stability of the policy rollouts and facilitate the training (see PLAL+M+S v. PLAL+M on Fig.~\ref{fig:pendulum_bench}).
The latter characteristic becomes even more important for tasks on more complex systems with chaotic dynamics (Sec.~\ref{sec:robot_exp}).

\subsection{Robotics systems} \label{sec:robot_exp}

Here, we consider tasks on the double pendulum and UR5 robotic arm.
These tasks are made very challenging by the highly non-linear and chaotic dynamics and the instability of the desired target configuration.

\begin{figure}
    \centering
    \includegraphics[width=0.45\linewidth]{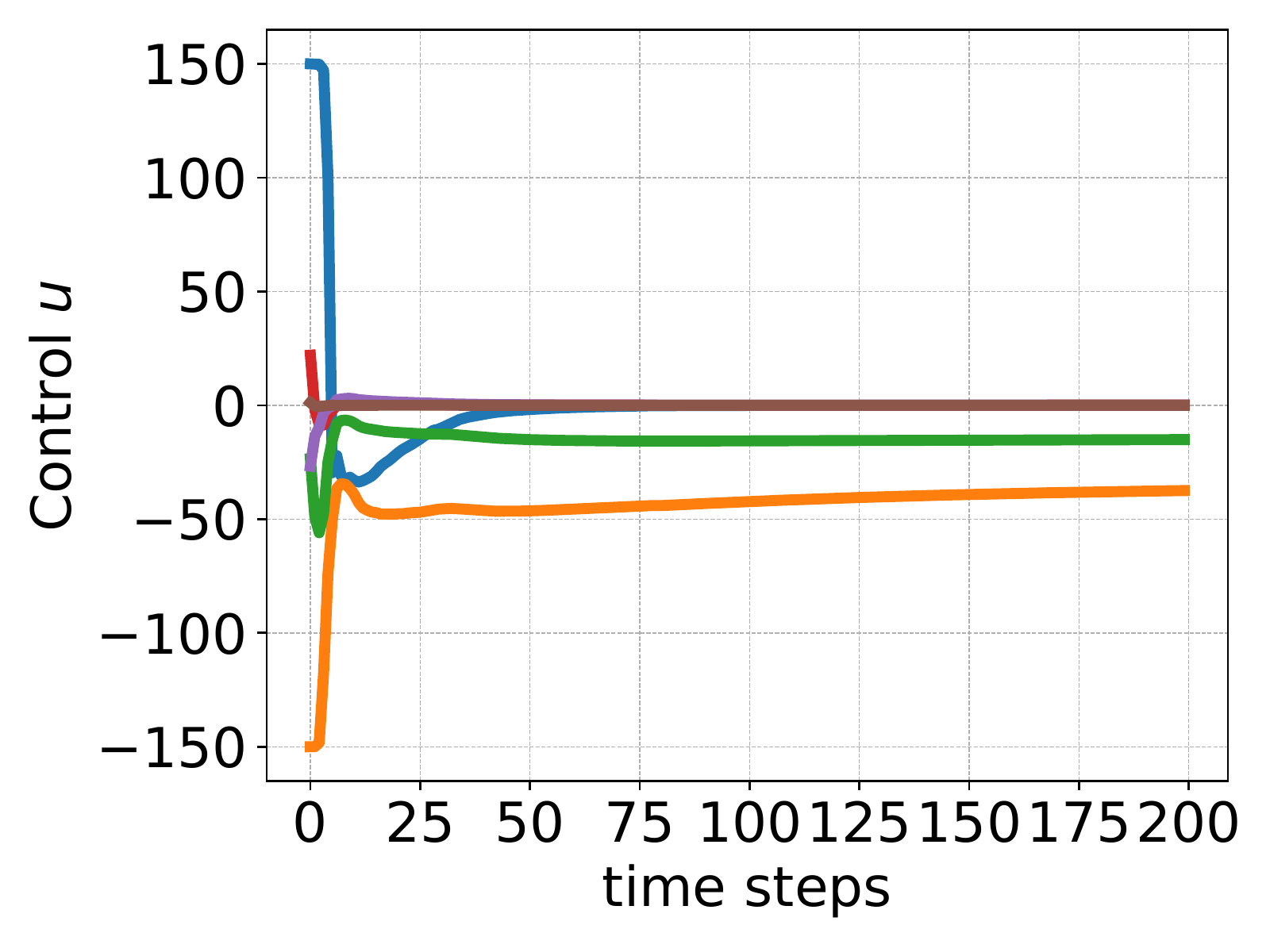}~
    \includegraphics[width=0.45\linewidth]{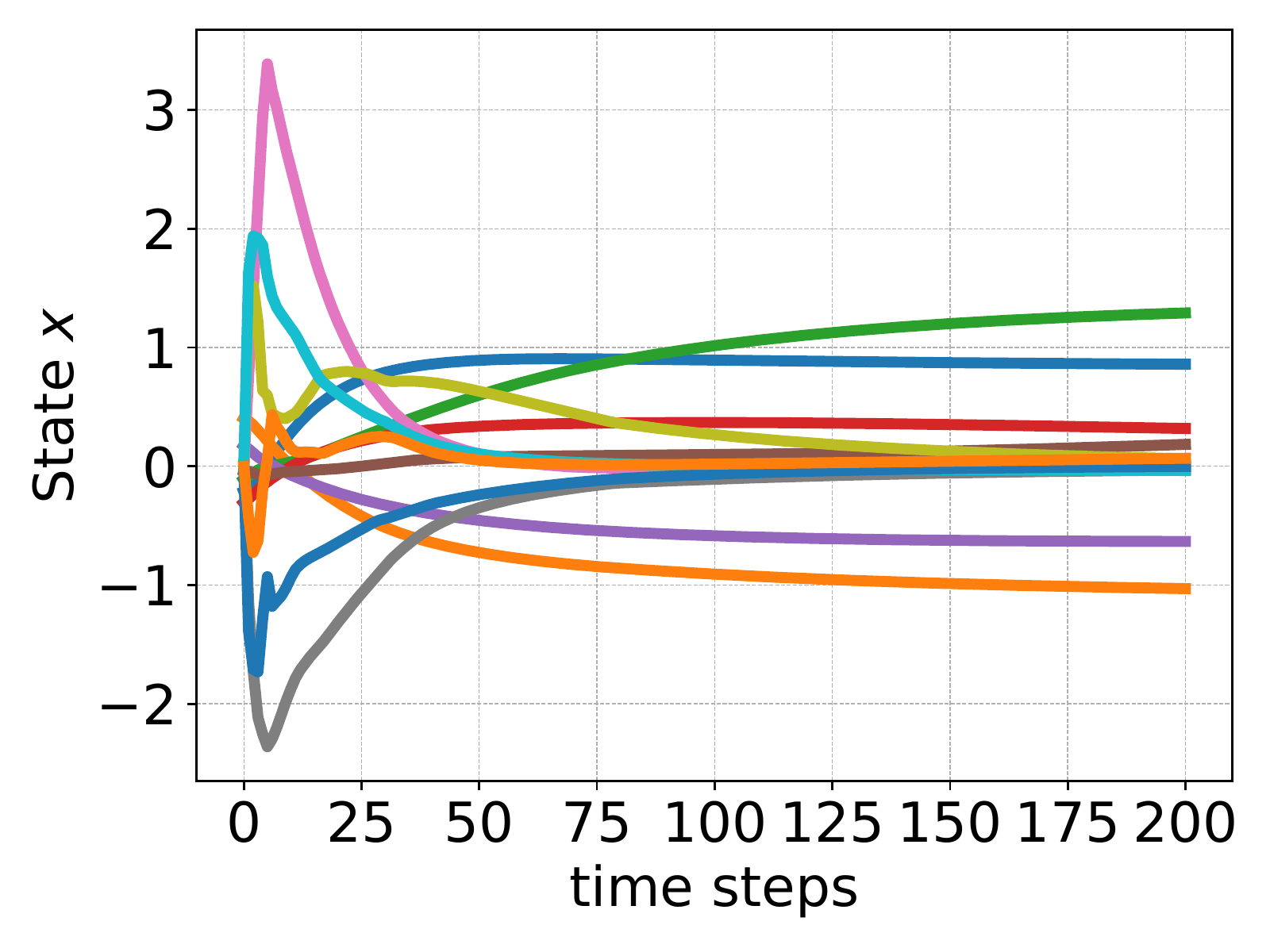}
    \includegraphics[width=0.45\linewidth]{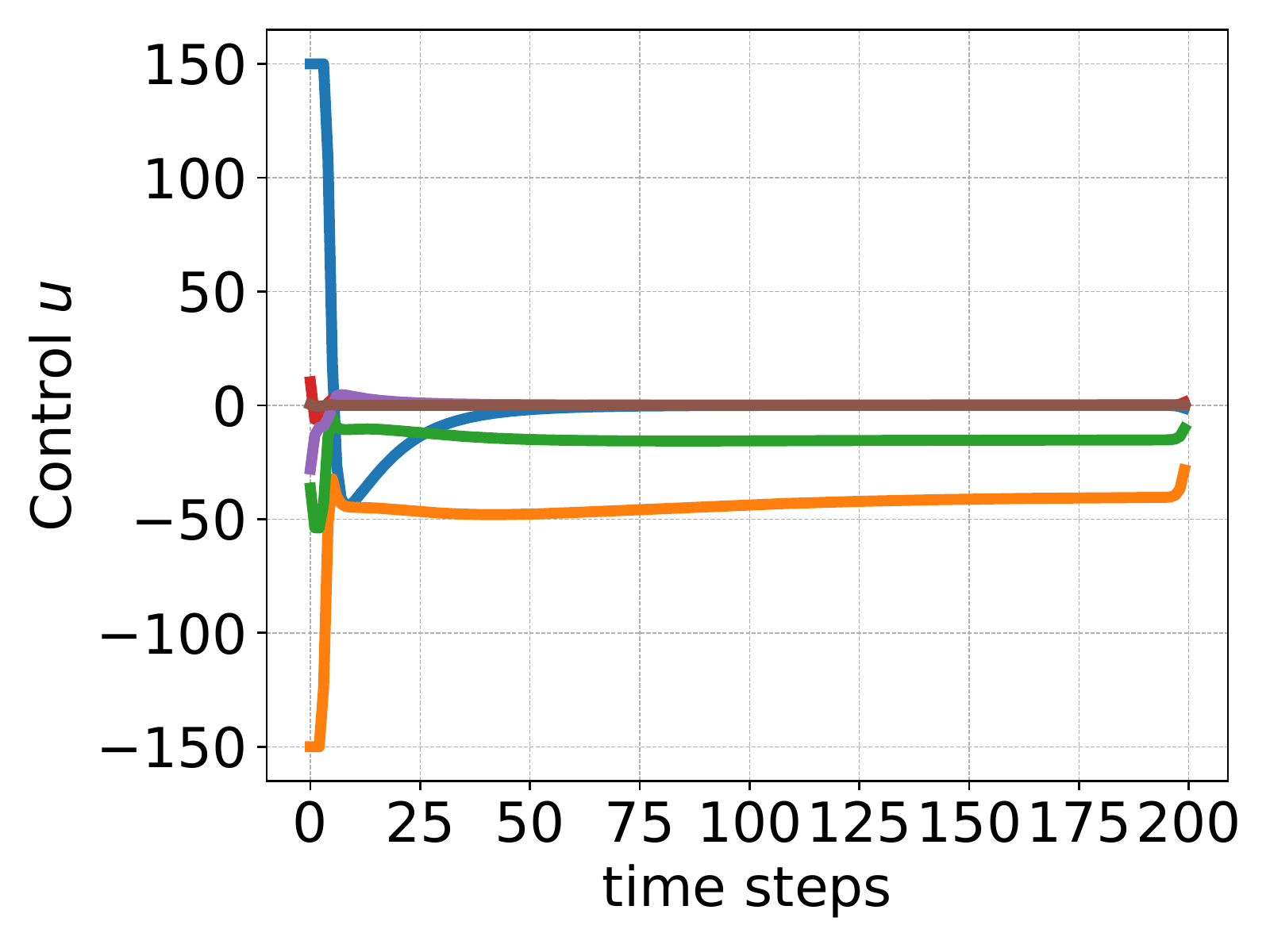}~
    \includegraphics[width=0.45\linewidth]{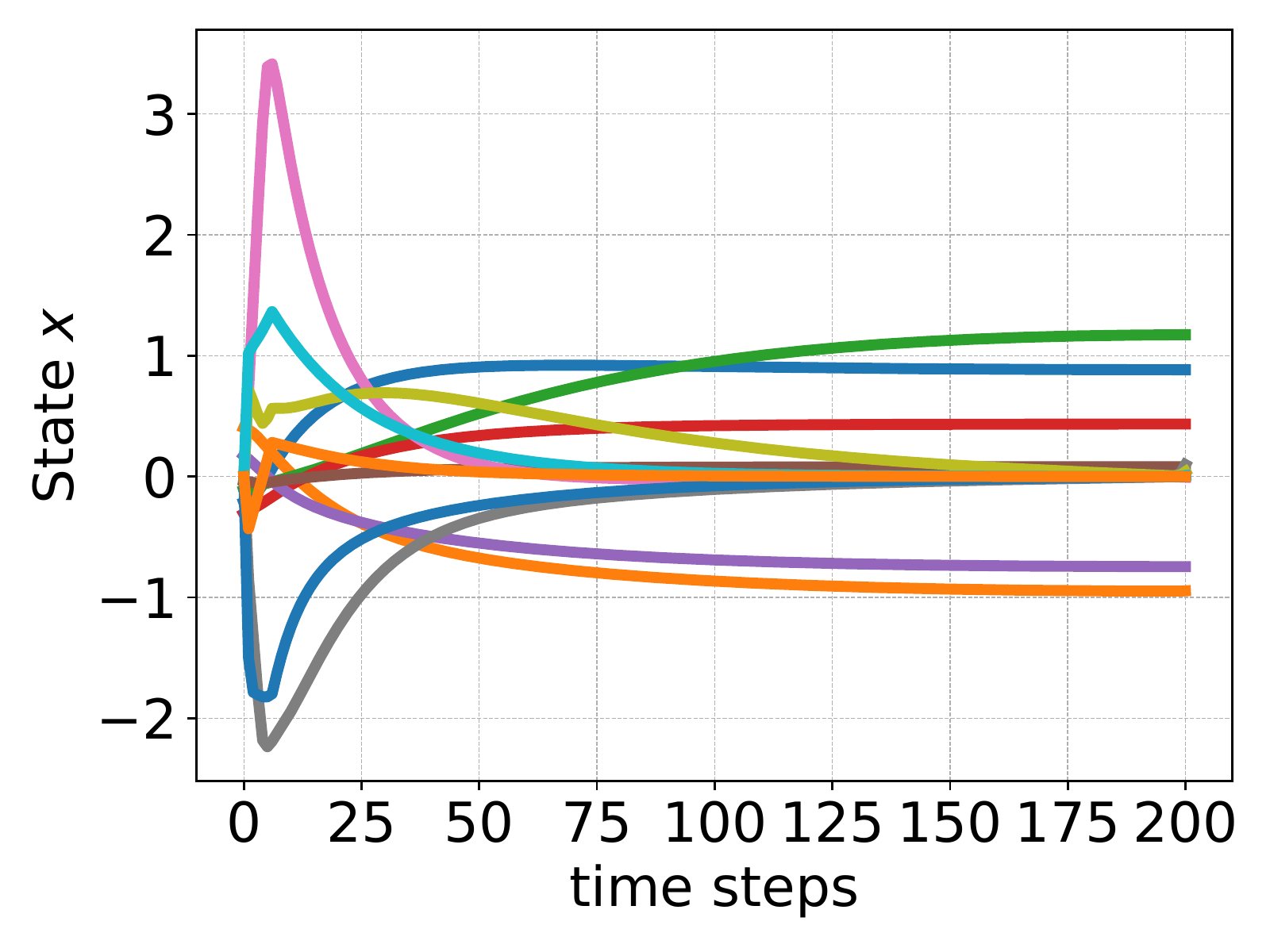}
    \includegraphics[width=0.45\linewidth]{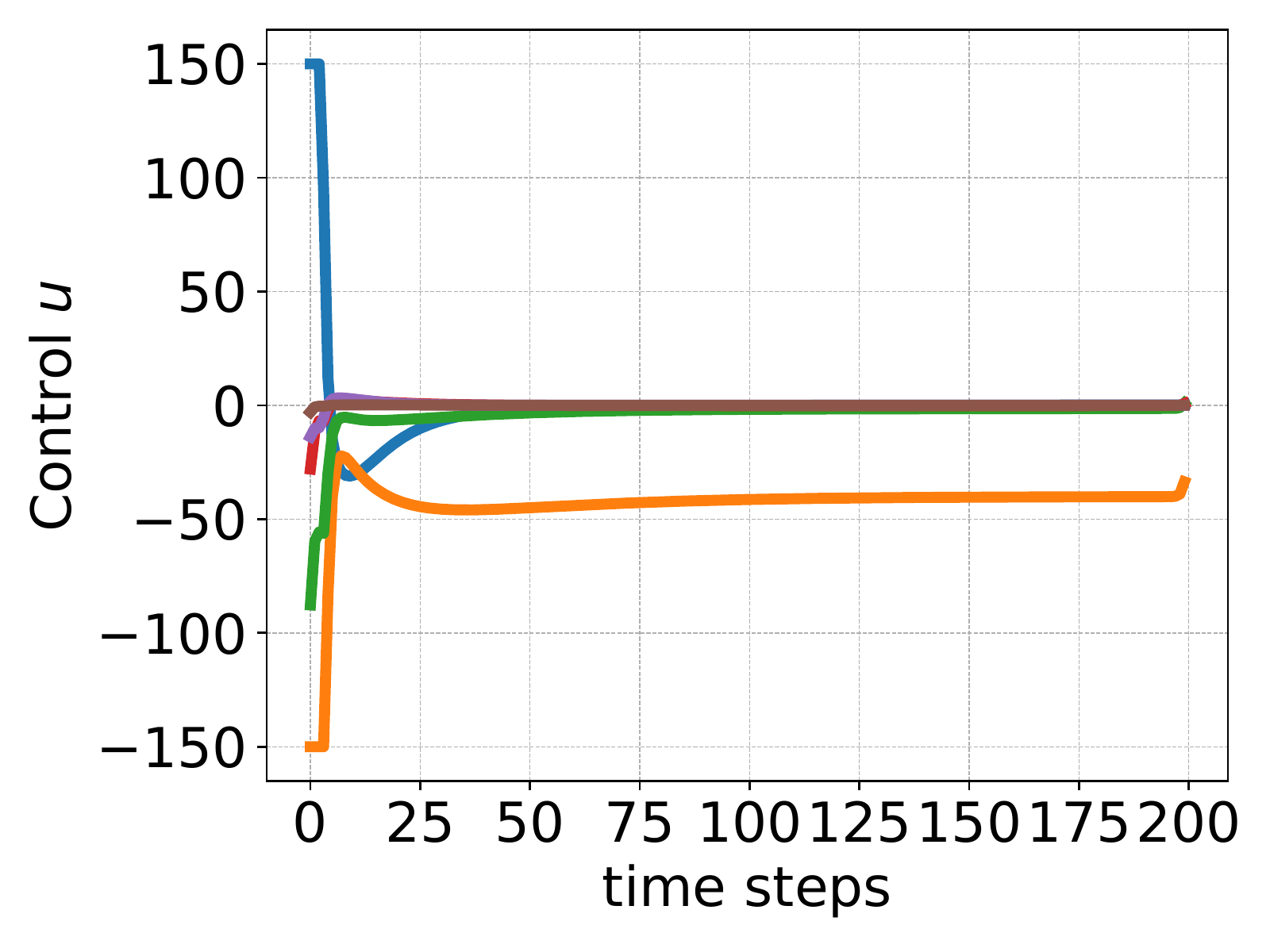}~
    \includegraphics[width=0.45\linewidth]{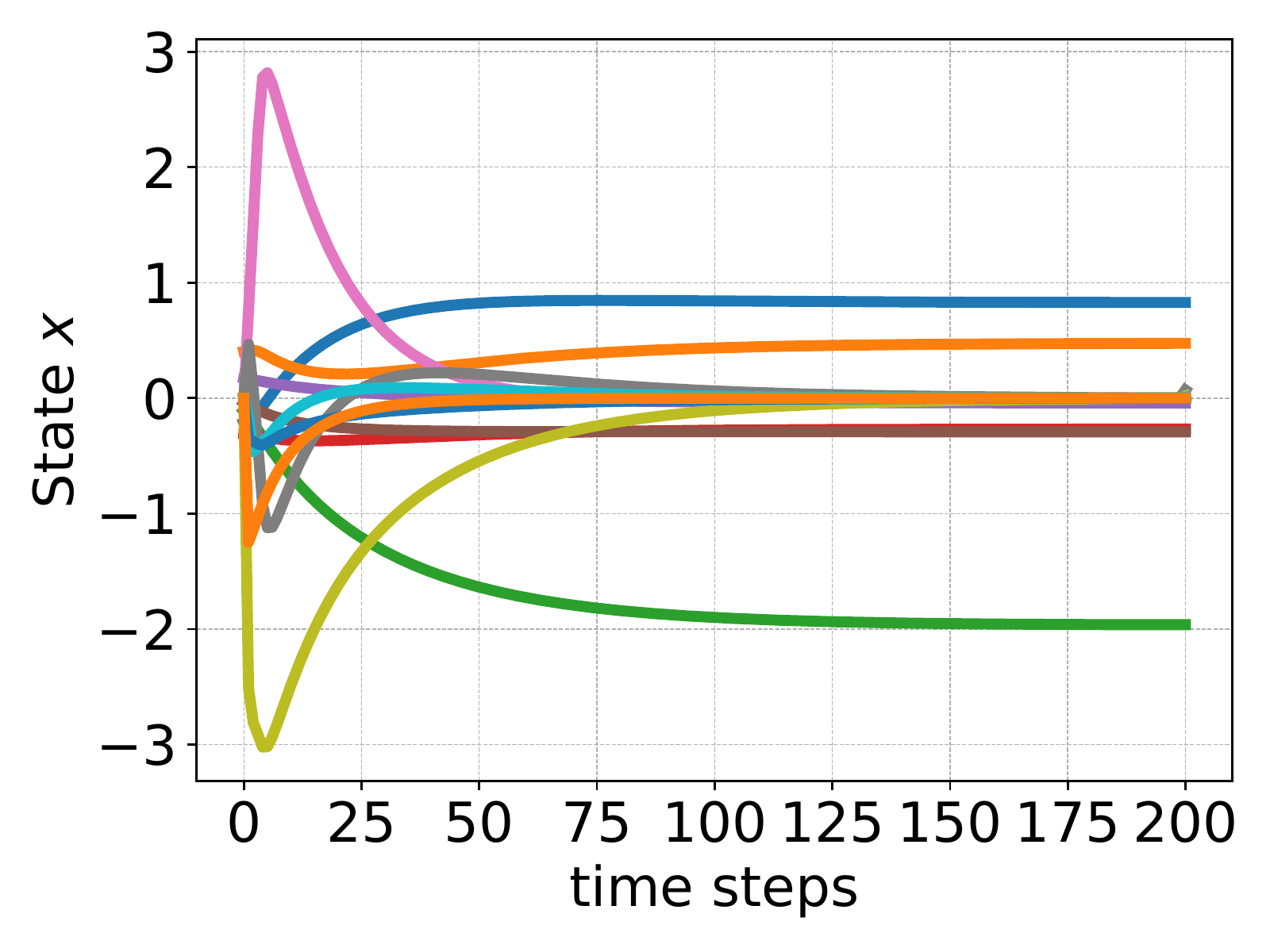}
    \caption{\small\textbf{Comparisons of controllers from TO and policy learning.} We represent the trajectories (\textbf{right}) and the corresponding control (\textbf{left}) on a testing sample of UR5. The $1$\textsuperscript{st} row corresponds to the learned policy while the $2$\textsuperscript{nd} and $3$\textsuperscript{rd} represent the trajectories obtained via TO when the solver is initialized respectively with and without the learned controller.}
    \label{fig:traj_pi_oc}
    \vspace{-0.4cm}
\end{figure}

\begin{figure}[h!]
    \centering
    \includegraphics[width=0.9\linewidth]{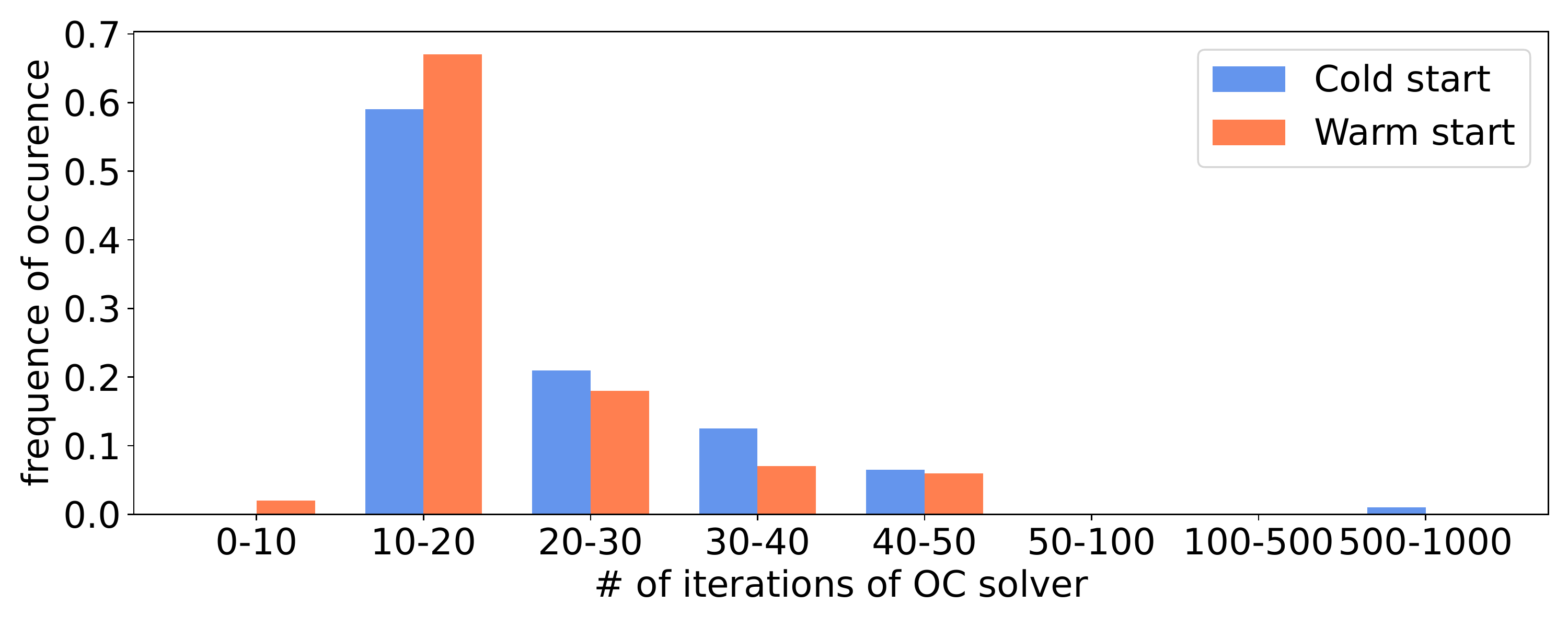}
    \caption{\small\textbf{Benefits of policy learning on TO.} We record the number of iterations required by the TO solver to reach a given precision on OCPs from the testing set.  We compare the cases where the solver is initialized with and without a rollout of the learned policy.}
    \label{fig:iters_warm_cold}
    \vspace{-0.65cm}
\end{figure}
As already noticed during experiments on the inverted pendulum, taking into account the physical constraints and exploiting the higher-order information contained in the feedback gains is crucial.
In the case of the double pendulum and UR5, the combination of these components even appears to be necessary to get converging algorithms (Fig.~\ref{fig:robot_bench}).
Due to the chaotic dynamics, for the PODS algorithm, even slight drifts from the policy w.r.t the TO controller induce very large deviations which, in turn, drive the policy towards saturation and cause the rollouts to diverge.
Including the physical constraints in the OC phase and regularizing the learning phase with the feedback gains allow to reduce the gap, and thus stabilize the training process.

Fig.~\ref{fig:robot_bench} (right) highlights the influence of the augmented Lagrangian parameter $\mu$  on the convergence of ADMM algorithms.
As described in \ref{sec:ablation}, multiple shooting already hinders the stability of the learning process, and when using it, we suspect the setting of $\mu$ to be even more crucial for the quality of the final solution (Fig.~\ref{fig:robot_bench}, right).

If we already demonstrated how policy learning can take advantage of TO, reciprocally, we observe that TO can also benefit from being warm-started by the learned policy (Fig.~\ref{fig:iters_warm_cold}).
In our case, doing so allows to systematically avoid worst case scenarii where the solver needs more than 100 steps.

Eventually, Fig.~\ref{fig:traj_pi_oc} demonstrates that warm-starting the solver with the policy also impacts the final control as it changes the local minima found by TO algorithms.
\section{Discussion, conclusion and future work} 
\label{sec:discussion}

In this paper, we have introduced a general framework to leverage the interplay between policy learning and trajectory optimization.
Our formulation results in two algorithmic variants which both exploit two constrained optimization techniques: a projected DDP method, and the alternating direction method of multipliers for establishing a strong consensus between the learned control policy and TO.
Both algorithms proceed in two alternating steps: solving an OCP and solving a supervised learning problem.
We have introduced enhancements for either part: constraints handling, stochastic Sobolev regularization and multiple shooting.
Although these algorithms achieve faster convergence towards more precise solutions when compared to classical RL algorithms, the experiments also highlight the need for advanced numerical trajectory optimization solvers.
Indeed, for challenging robotic tasks, constraint handling and higher order information appear to be necessary to stabilize the training of the policy.
The experiments and the ablation study also reveal the interest in our proposed enhancements: stochastic Sobolev learning enhanced sample efficiency and the stability of learned policy rollouts, and multiple shooting eliminates drift in the optimized trajectories.

A possible enhancement is the convergence speed of the method towards a consensus between the OCP and the supervised learning loop.
It is known that Augmented Lagrangian methods (which includes ADMM) can reach much higher performance by having a strategy for updating the penalty parameter $(\mu_t)$. One such strategy in the literature is that of the bound-constrained Lagrangian (BCL) method of~\cite{connGloballyConvergentAugmented1991}, which could be a direction for further enhancements.

In this work, we assumed the samples $(\beta^{(i)})_i$ used for policy optimization are fixed; they are never re-sampled (which is how stochastic algorithms such as SGD work) nor are new samples ever added. This is the domain of constrained stochastic optimization; one of the difficulties here is updating the Lagrange multipliers in a way where optimization progress is not lost at every sampling. Such an extension could allow active sampling of the initial condition $x^0$ or other parameters so to explore the state space and learn policies more effectively.

We could then explore extensions towards setups closer to GPS~\cite{levine2016end} by considering partially unknown system dynamics, with limited knowledge of some physical parameters (\textit{e.g.} friction coefficients) -- in order to take advantage of physical models once again.
This would require embedding a system identification step in the loop which would also benefit from differentiable simulation techniques.

\begin{footnotesize}
\section*{Acknowledgments}
This work was supported in part by L'Agence d'Innovation Défense, the HPC resources from GENCI-IDRIS(Grant AD011012215), the French government under management of Agence Nationale de la Recherche as part of the "Investissements d'avenir" program, reference ANR-19-P3IA-0001 (PRAIRIE 3IA Institute), the AGIMUS project, funded by the European Union under GA no.101070165 and Louis Vuitton ENS Chair on Artificial Intelligence. 
\end{footnotesize}

\clearpage
\balance
\bibliographystyle{ieeetr}
\bibliography{bibliography}

\end{document}